\documentclass[10pt,twocolumn,letterpaper]{article}
\usepackage{cvpr}
\usepackage{times}
\usepackage{epsfig}
\usepackage{graphicx}
\usepackage{amsmath}
\usepackage{amssymb}
\usepackage{booktabs}
\usepackage{url}
\usepackage{epstopdf}
\usepackage{cite}
\usepackage{gensymb}
\usepackage{url}
\usepackage[bottom]{footmisc}
\usepackage{enumitem}
\usepackage{caption}
\usepackage{float}
\usepackage{subcaption}
\usepackage{multirow} 
\usepackage{color}
\usepackage{breqn}
\usepackage{bbm}
\usepackage{tabularx}
\graphicspath{{./fig/}}
\def\Vec#1{{\boldsymbol{#1}}}
\def\Mat#1{{\boldsymbol{#1}}}
\renewcommand{\Vec}[1]{\boldsymbol{\mathbf{#1}}}
\renewcommand{\Mat}[1]{\boldsymbol{\mathbf{#1}}}

\usepackage[pagebackref=true,breaklinks=true,letterpaper=true,colorlinks,bookmarks=false]{hyperref}

\cvprfinalcopy 

\def\cvprPaperID{1230} 

\ifcvprfinal\fi

\begin{document}

\title{Disentangling 3D Pose in A Dendritic CNN \\for Unconstrained 2D Face Alignment}

\author{  Amit Kumar \hspace{15pt} Rama Chellappa\\
Department of Electrical and Computer Engineering, CFAR and UMIACS\\
     University of Maryland-College Park,USA \\
{\tt\small akumar14@umiacs.umd.edu, rama@umiacs.umd.edu}
}

\maketitle


\begin{abstract}
Heatmap regression has been used for landmark localization for quite a while now. Most of the methods use a very deep stack of bottleneck modules for heatmap classification stage, followed by heatmap regression to extract the keypoints. In this paper, we present a single dendritic CNN, termed as Pose Conditioned Dendritic Convolution Neural Network (PCD-CNN), where a classification network is followed by a second and modular classification network, trained in an end to end fashion to obtain accurate landmark points. Following a Bayesian formulation, we disentangle the 3D pose of a face image explicitly by conditioning the landmark estimation on pose, making it different from multi-tasking approaches. Extensive experimentation shows that conditioning on pose reduces the localization error by making it agnostic to face pose. The proposed model can be extended to yield variable number of landmark points and hence broadening its applicability to other datasets. Instead of increasing depth or width of the network, we train the CNN efficiently with Mask-Softmax Loss and hard sample mining to achieve upto $15\%$ reduction in error compared to state-of-the-art methods for extreme and medium pose face images from challenging datasets including AFLW, AFW, COFW and IBUG. 
\end{abstract}

\section{Introduction}
\begin{figure}[t]
\centering
\includegraphics[height=3cm,width=0.5\textwidth]{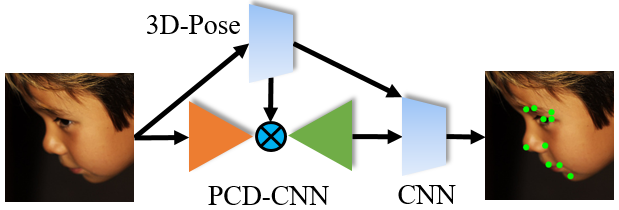}
\caption{\small A bird's eye view of the proposed method. Dendritic CNN is explicitly conditioned on 3D pose. A generic CNN is used for auxiliary tasks such as fine-grained localization or occlusion detection.}
\label{fig:cover}
\end{figure}
Face alignment or facial landmark estimation is the task of estimating keypoints such as eye-corners, mouth corners etc. on a face image. As shown in \cite{bansal2017dosanddonts}, accurate face alignment improves the performance of a face verification system \cite{DBLP:conf/wacv/XuZAC16,DBLP:conf/icpr/XuZAC16,DBLP:conf/wacv/BodlaZXCCC17}, as well as other applications such as 3D face modelling, face animation etc. Currently, face alignment is dominated by regression-based approaches which yield a fixed number of points. Explicit Shape Regression (ESR)\cite{DBLP:journals/ijcv/CaoWWS14} and Supervised Descent Method (SDM)\cite{XiongD13} have addressed the problem of face alignment for faces in medium pose. To achieve sub-pixel accuracy on such face images, coarse to fine approaches have also been proposed in the literature\cite{CFAN,Zhu_2015_CVPR,DBLP:journals/corr/KumarRPC16}. It is evident that such methods perform  poorly on face images with extreme pose, expression and lighting mainly because they are dependent on bounding box and mean face shape intializations. On the other hand, Convolutional Neural Networks (CNNs) have achieved breakthroughs in many vision tasks including the task of keypoints estimation\cite{Newell2016}. Lately, researchers have used heatmap regression extensively for the task of face alignment and pose estimation using an Encoder-Decoder architecture in the form of Convolution-Deconvolution Networks\cite{DBLP:journals/corr/ChenPKMY14}. Most of the approaches in the literature perform heatmap classification followed by regression\cite{Bulat2016,7961778,Bulat_2017_ICCV,BMVC2016_86}. In this paper, we propose the Pose Conditioned Dendritic Convolution Neural Network (PCD-CNN); which models the dendritic  structure of facial landmarks using a single CNN (see Figure \ref{fig:cover}). 

\textbf{Shape constraint}: Methods such as ESR\cite{DBLP:journals/ijcv/CaoWWS14} and SDM\cite{XiongD13} impose the shape constraint by jointly regressing over all the points. Such a shape constraint cannot be applied to a profile face as a consequence of extreme pose leading to a variable number of points. Tree structured part models (TSPM)\cite{AFW_dataset_CVPR2012} by Zhu et al. had two major limitations associated with it; namely pre-determined models and slower run-time. With an intent to solve these, we propose a tree structure model in a single Dendritic CNN (PCD-CNN), which is able to capture the shape constraint in a deep learning framework. 

\textbf{Pose}: Works such as Hyperface\cite{DBLP:journals/corr/RanjanPC16} and TCDCN\cite{DBLP:conf/eccv/ZhangLLT14} have used 3D pose in a multitask framework and demonstrated that learning pose and keypoints jointly using a deep network improves the performance of both tasks. 
However, in contrast to multi-tasking approaches, we condition the landmark estimates on the head pose, following a Bayesian formulation and demonstrate the effectiveness of the proposed approach through extensive experiments. We wish to point out that our primary goal is not to predict the head pose, instead, use 3D head pose to condition the landmark points. This makes our work different from multitask approaches.

\textbf{Speed-vs-Accuracy}: We observe that systems which process images at real time, such as \cite{Bhagavatula_2017_ICCV,large-pose-face-alignment-via-cnn-based-dense-3d-model-fitting} have higher error rate as opposed to cascade methods which are accurate but slow. Researchers have proposed many different network architectures like Hourglass\cite{Newell2016}, Binarized CNN (based on hourglass)\cite{Bulat_2017_ICCV} in order to achieve accuracy in keypoints estimation. Although, such methods are fully convolutional , they suffer from slower run time as a result of cascaded deep bottleneck modules which perform a large number of FLOPs during test time. The proposed PCD-CNN works at the same scale as the input image and thus reduces the extrapolation errors. PCD-CNN is fully convolutional with fewer parameters and is capable of processing images almost at real time speed ($20$FPS). Limited generalizability as a consequence of smaller number of parameters is tackled by efficiently training the network using Mask-Softmax loss and difficult sample mining.

\textbf{Generalizability}: Methods for domain-limited face images have been developed, mostly following the cascade regression approach. \cite{10.1109/ICCV.2013.191,Zhang_2016_CVPR,Wu_2017_CVPR} have been shown to work well for faces under extreme external object occlusion. On the other hand, \cite{Tzimiropoulos_2015_CVPR,trigeorgis2016mnemonic,6909635,7299048,Zhu_2015_CVPR,DBLP:conf/cvpr/RenCW014} achieved satisfactory results on the 300W\cite{6755925} dataset which contains images in medium pose with almost no occlusion.\cite{7961750,Zhu_2016_CVPR,pifa} have demonstrated their effectiveness for extreme pose datasets with a limited number of fiducial points. However, they do not generalize very well to other datasets. We show that by a small increase in the number of parameters, PCD-CNN can be extended to most of the publicly available datasets including 300W, COFW, AFLW and AFW yielding variable number of points depending on the protocol.  
\begin{figure*}[h]
    \centering
    \begin{subfigure}[]{0.25\textwidth}
        \centering
        \includegraphics[height=5.5cm,width=13.75cm]{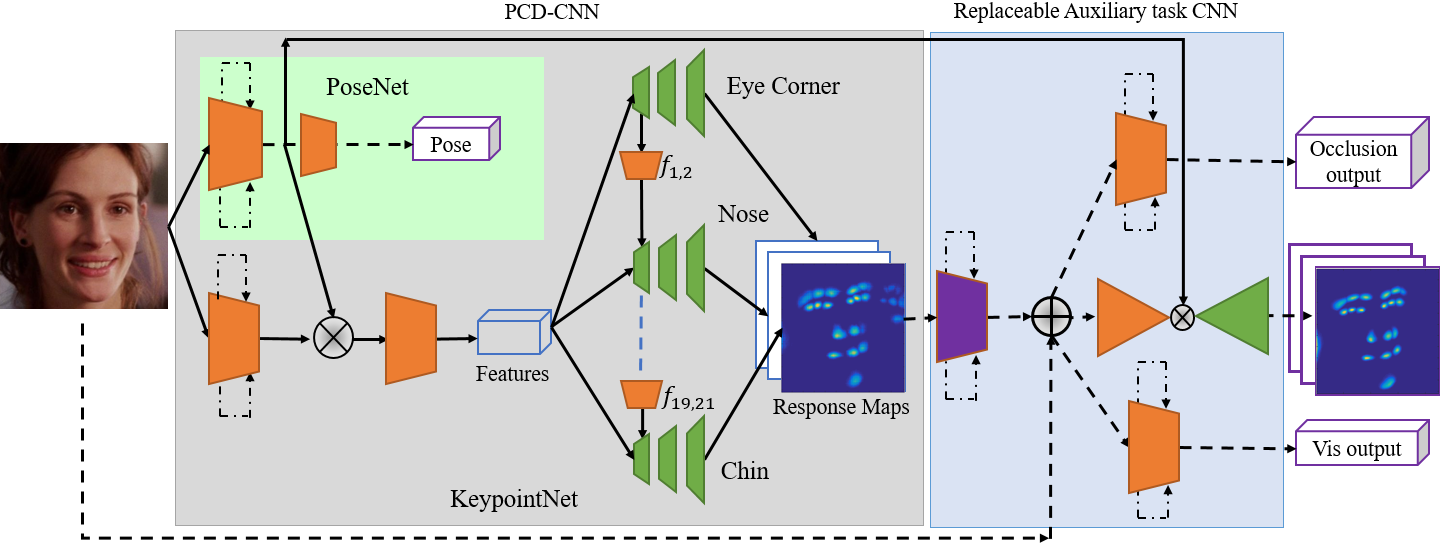}
        \caption{}
        \label{fig:outline_a}
    \end{subfigure}  \hfill  
    \begin{subfigure}[]{0.2\textwidth}
        \centering
        \includegraphics[width=1\linewidth]{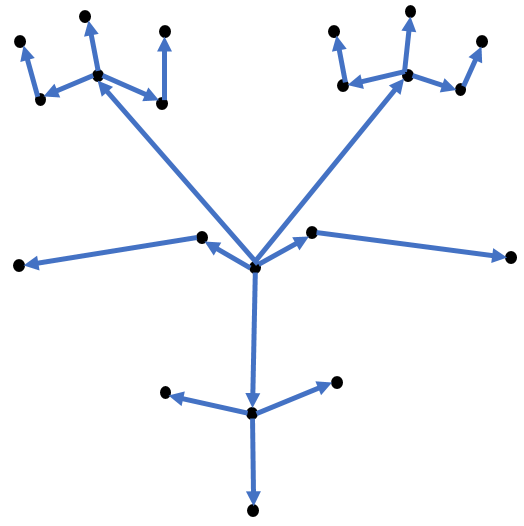}
        \caption{}
        \label{fig:outline_b}
    \end{subfigure}
    \caption{\small (a) Details of the proposed method. The dotted lines on top of convolution layers denote residual connections. Dendritic KeypointNet is conditioned on PoseNet. The network inside the grey box represents the proposed PCD-CNN, whereas the second network inside the blue box is modular and can be replaced for an auxiliary task. A conv-deconv network for finer localization is used alongside these auxiliary networks. (b) Proposed dendritic structure of facial landmark points for effective information sharing among landmark points. The nodes of the dendritic structure are the outputs of deconvolutions while the edges between nodes $i$ and $j$ are modeled by convolution functions $f_{ij}$. For the architecture of deconvolution network refer to Figure \ref{fig:deconv}. }
     \label{fig:outline}
\end{figure*}


Following the above discussion, the main contributions of this paper can be listed as: 
\begin{itemize}[noitemsep,nolistsep]
\item We propose the Pose Disentangled Dendritic CNN for unconstrained 2D face alignment, where the shape constraint is imposed by the dendritic structure of facial landmarks. The proposed method uses classification followed by classification approach as opposed to classification followed by regression. The second auxiliary network is modular and can be designed for fine grained localization or any other auxiliary tasks. Figure \ref{fig:outline} shows the overall structure of PCD-CNN. 
\item The proposed method disentangles the head pose using a Bayesian framework and experimentally demonstrates that conditioning on 3D head pose improves the localization performance. The proposed method processes images at real-time speed producing accurate results. 
\item With a recursive extension, the proposed method can be extended to datasets with arbitrarily different number of points and different auxiliary tasks. 
\item As a by-product, the network outputs pose estimates of the face image where we achieve close to state-of-the-art result on pose estimation on the AFW dataset. In another experiment, the auxiliary classification network is trained for occlusion detection where we obtain state-of-the-art result for occlusion detection on COFW dataset. 
\end{itemize}
\section{Prior Work}
We briefly review prior work in the area of keypoint localization under the following two categories: Deep Learning-based and Hand crafted features-based methods. 

Parametric part-based models such as Active Appearance Models (AAMs)\cite{AAM} and Constrained Local Models\cite{Cristinacce:2008:AFL:1385702.1385969} are statistical methods which perform keypoint detection by maximizing the confidence of part locations in a given input image using \textit{\textbf{handcrafted features}} such as SIFT and HOG.  The tree structure part model (TSPM) proposed in \cite{AFW_dataset_CVPR2012} used deformable part-based model for simultaneous detection, pose estimation and landmark localization of face images modeling the face shape in a mixture of trees model. Later, \cite{Asthana:2013:RDR:2514950.2516059} proposed learning a dictionary of probability response maps followed by linear regression in a Constrained Local Model (CLM) framework. Early cascade regression-based methods such as \cite{DBLP:journals/ijcv/CaoWWS14,Zhu_2015_CVPR,trigeorgis2016mnemonic,akshay_wild,6909635,XiongD13,6619290} also used hand crafted features such as SIFT to capture appearance of the face image. The major drawback of regression-based methods is their inability to learn models for unconstrained faces in extreme pose.

\textbf{\textit{Deep learning}}-based methods have achieved breakthroughs in a variety of vision tasks including landmark localization. One of the earliest works was done in\cite{Sun:2013:DCN:2514950.2516090,DBLP:journals/corr/KumarRPC16} where a cascade of deep models was learnt for fiducial detection. 3DDFA\cite{DBLP:journals/corr/ZhuLLSL15} modeled the depth of the face image in a Z-buffer, after which a dense 3D face model was fitted to the image via CNNs. 
Pose Invariant Face Alignment (PIFA)\cite{pifa} by Jourabloo et al. predicted the coefficients of 3D to 2D projection matrix via deep cascade regressors. 
\cite{Bhagavatula_2017_ICCV} used 3D spatial transformer networks to capture 3D to 2D projection. \cite{jackson2017vrn,dense-face-alignment,pose-invariant-face-alignment-with-a-single-cnn} extended \cite{pifa} by using CNNs to directly learn the dense 3D coordinates.  
The proposed method has a dendritic structure which looks at the global appearance of the image while the local interactions are captured by pose conditioned convolutions. PCD-CNN does not assume that all the keypoints are visible and the interactions between keypoints are learned. PCD-CNN is entirely based on 2D images, which captures the 3D information by conditioning on 3D head pose.    

Formulating keypoint estimation as the per-pixel labeling task, Hourglass networks\cite{Newell2016} and Structured feature learning\cite{chu2016structure} were proposed. Hourglass networks use a stack of $8$ very deep hourglass modules and hence, even though based entirely on convolution can process only $8$-$10$ frames per second. \cite{chu2016structure} implemented message passing between keypoints, however was able to process images at lower resolution due to large number of parameters. PCD-CNN models the dendritic structure in branched deconvolution networks where each network is implemented in Squeezenet\cite{SqueezeNet} fashion and hence has fewer parameters, contributing to real-time operation at full image scale.  

In the next few sections, we describe Pose Conditioned Dendritic-CNN in detail where we discuss the different concepts introduced, and then present ablative studies to arrive at the desired architecture.
\section{Pose Conditioned Dendritic CNN}

The task of keypoint detection is to estimate the 2D coordinates of, say \textit{N} landmark points, given a face image. Observing the effectiveness of deep networks for a variety of vision tasks, we present a single end-to-end trainable deep neural network for landmark localization. 

\textbf{Conditioning on 3D pose}: Keypoints are susceptible to variations in external factors such as emotion, occlusion and intrinsic face shape. On the other hand, 3D pose is fairly stable to them and can be estimated directly from 2D image\cite{7961750}. Reasonably accurate 2D keypoint coordinates can be also inferred given 3D pose and a generic 3D model of a human face. However, the converse problem of estimating 3D pose from 2D keypoints is ill posed. Therefore, we make use of the probabilistic formulation over the variables including the image $\Mat{I}\in\mathbb{R}^{w\times h\times 3}$ of height $h$ and width $w$, $3D$ head pose denoted by $\Mat{P}\in \mathbb{R}^{3}$, $2D$ keypoints $\Mat{C}\in\mathbb{R}^{N\times2}$, where $N$ is the number of keypoints. Following the natural hierarchy between the two tasks, the joint and the conditional probabilities can be written as:
\begin{equation}
\label{eq00}
p(\Mat{C},\Mat{P},\Mat{I}) = p(\Mat{C}|\Mat{P},\Mat{I}) p(\Mat{P}|\Mat{I}) p(\Mat{I}) 
\end{equation}
\begin{align}
\label{eq01}
p(\Mat{C},\Mat{P}|\Mat{I}) &= \frac{p(\Mat{C},\Mat{P},\Mat{I})}{p(\Mat{I})} \nonumber \\
&= \underbrace{p(\Mat{P}|\Mat{I})}_{\text{CNN}} . \underbrace{p(\Mat{C}|\Mat{P},\Mat{I})}_{\text{PCD-CNN}}
\end{align}
We implement the first factor with an image-based CNN learned to predict the 3D pose of the face image. The second factor is implemented through a ConvNet and multiple DeconvNets arranged in a dendritic structure. The convolution network maps the image to lower dimension, after which the outputs of several deconvolution networks are stacked to form the keypoint-heatmap. The models are tied together by element-wise product (as (\ref{eq00}) and (\ref{eq01})) to condition the measurement of 2D coordinates on 3D pose. We choose element-wise product as the operation to condition on the head pose as keypoint heatmaps can be interpreted as probability distribution over the keypoints. 
The visibility of each keypoint is learnt implicitly as the invisible points are labeled as background. 

\textbf{Multi-tasking-vs-Conditioning}: In a multi-tasking method such as\cite{7961750}, several tasks are learnt synergetically and backpropagation impacts all the tasks. On the other hand, in the proposed PCD-CNN, the error gradients backpropagated from keypoint network affect both, keypoint network and pose network; however, the pose network affects the keypoint network only during the forward pass. In other words, multi-tasking approaches try to model the joint distribution $p(\Mat{C},\Mat{P}|\Mat{I})$ , whereas the proposed approach explicitly models the decomposed form $p(\Mat{P}|\Mat{I})p(\Mat{C}|\Mat{P},\Mat{I})$ by learning the individual factors.

\textbf{Proposed Pose Conditioned Dendritic CNN} :
To capture the structural relationship between different keypoints, we propose the dendritic structure of facial landmarks as shown in figure \ref{fig:outline_b} where the nose tip is assumed to be the root node. Such a structure is feasible even in faces with extreme pose. Following this, the keypoint network is modeled with a single CNN in a tree structure composed of convolution and deconvolution layers. The pairwise relationships between different keypoints are modeled via specialized functions, $f_{i,j}$, which are implemented through convolutions and are analogous to the spring weights in the spring-weight model of Deformable Part Models\cite{Felzenszwalb:2010:ODD:1850486.1850574}. A low confidence of a particular keypoint is reinforced when the response of $f_{i,j}$ corresponding to the adjacent node is added. With experimental justifications we show that such a deformable tree model outperforms the recently published works \cite{ large-pose-face-alignment-via-cnn-based-dense-3d-model-fitting,Bhagavatula_2017_ICCV,pose-invariant-face-alignment-with-a-single-cnn,dense-face-alignment} which use 3D models and 3D spatial transformer networks to supplement keypoint detection models. Figure \ref{fig:outline} shows the overall architecture of the proposed PCD-CNN and the proposed dendritic structure of the facial landmarks.

Instead of going deeper or wider\cite{Newell2016,Bulat_2017_ICCV} with deep networks, we base our work on the Squeezenet-11\cite{SqueezeNet} architecture, attributing to its capability to maintain performance with fewer parameters. 
We use two Squeezenet-11 networks; one for pose and other for keypoints, named as -PoseNet and KeypointNet respectively, as shown in Fig \ref{fig:outline_a}. Convolutions are performed on the $pool_{8}$ activation maps of the PoseNet, the response of which is then multiplied element-wise to the response maps of $pool_{8}$ layers of the KeypointNet. Each convolution layer is followed by ReLU non-linearity and batch normalization. In table \ref{pose_condition_table}, we show that keypoint localization error reduces when conditioned on 3D head pose. 

\begin{figure}[t]
\centering
\includegraphics[height=3.65cm,width=6cm]{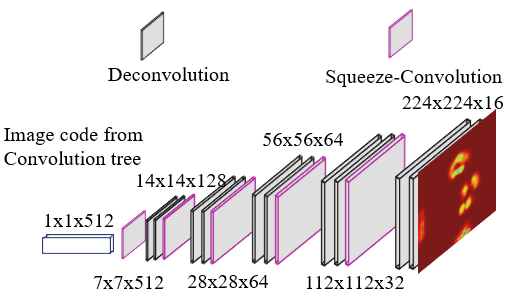}
\caption{\small Detailed description of a single Squeezenet-DeconvNet network. Note the fewer number of deconvolution filters. Each deconvolution network is identical to the one shown above.}
\label{fig:deconv}
\end{figure}

The design of deconvolution network is non-trivial. To maintain the same property as of SqueezeNet, we first upsample the feature maps using parametrized strided convolutions and then squeeze the output features maps using $1\mathrm{x}1$ convolutions.  We call this network as Squeezenet-DeconvNet. Figure \ref{fig:deconv} shows the detailed architecture of the Squeezenet-DeconvNet. Since, each keypoint in the proposed network is modeled by a separate Squeezenet-DeconvNet, it alleviates the need for large number of deconvolution parameters ($256$ and $512$ $3\times3$ in Hourglass networks). In fact, in the practical version of PCD-CNN, there are only 32 and 16 deconvolution filters which results in the design of networks, which are small enough to fit in a single GPU. The design of networks with fewer filters is motivated by real-time processing consideration. With experiments we show that disentangling the pose by conditioning on it, reinforces the learning of the proposed PCD-CNN with fewer parameters (Table \ref{pose_condition_table}). 

\begin{table}[]
\begin{subtable}[h]{0.5\textwidth}
\centering
\begin{tabular}{|l|l|}
\hline
\textbf{Method} & \textbf{Normalised Error} \\
\hline \hline 
Without pose conditioning & 3.45 \\
With pose conditioning & 2.85 \\
\hline
\end{tabular}
\caption{}
\label{pose_condition_table}
\end{subtable}
\hfill \hfill
\begin{subtable}[h]{0.5\textwidth}
\centering
\begin{tabular}{|l|l|}
\hline
\textbf{Method} & \textbf{Normalised Error} \\
\hline \hline 
Classification+Regression & 3.93 \\
Classification+Classification & 3.09 \\
\hline
\end{tabular}
\caption{}
\label{Regression_v_classification}
\end{subtable}
\hfill \hfill
\begin{subtable}[h]{0.5\textwidth}
\centering
\begin{tabular}{|l|l|}
\hline
\textbf{Method} & \textbf{Normalised Error} \\
\hline \hline 
Softmax & 4.56 \\
Using Mask-Softmax & 2.85 \\
\hline
\end{tabular}
\caption{}
\label{mask_softmax_table}
\end{subtable}
\caption{\small Root mean square error normalized by bounding box size, calculated on the AFLW validation set following the PIFA protocol. (a) With and without conditioning on pose. (b) Comparison showing that PCD-CNN when followed by another classification stage results in lower localization error compared to classification followed by regression. Note that conditioning on pose is not used in both the cases above for fair comparison. (c) Comparison indicating the effect of using Mask-softmax over Softmax }
\end{table}


In order to obtain fine grained localization results, we concatenate to the input data, a learned function of the predicted probabilities (represented as purple box in Figure \ref{fig:outline_a}) and pass them through the second Squeezenet based conv-deconv network. This function is modeled by a residual unit with $1\times1$ and $3\times3$ filters, which are learned end-to-end with the second classification network (while keeping the weights PCD-CNN frozen). For experimental purposes, we replace the second conv-deconv by another regression network designed along the lines of GoogleNet\cite{c2}. Table \ref{Regression_v_classification} shows a comparison between two stage classification approach versus classifcation followed by regression approaches \cite{recdec}.    

One of the goals of this work is to generalize the facial landmark detection to other datasets in order to broaden its applicability. A trivial extension would be to increase the number of deconvolution branches, which however is infeasible due to limited GPU memory. However, PCD-CNN can be extended to yield more landmark points arranged in different configurations. In figure \ref{fig:tree_extension} we show the proposed tree structures for COFW and 300W datasets with $29$ and $68$ landmark points respectively. Keeping the basic \textbf{Dendritic Structure of Parts} intact, first the number of output response maps in the last deconvolution layer are increased and then network slicing is performed to produce the desired number of keypoints. For instance, the output of the deconvolution network for eye-center is sliced to produce four outputs as required by the 300W dataset. Depending on the dataset, the second network can be replaced to perform auxiliary tasks resulting in a modular architecture; for instance in the case of COFW dataset we replace the second conv-deconv network with another Squeezenet network to detect occlusion. We direct the readers to the supplementary material for more details on network surgery and a magnified view of figures \ref{fig:outline_b} and \ref{fig:tree_extension}.

\begin{figure}[htp!]
\centering
\includegraphics[scale=0.21]{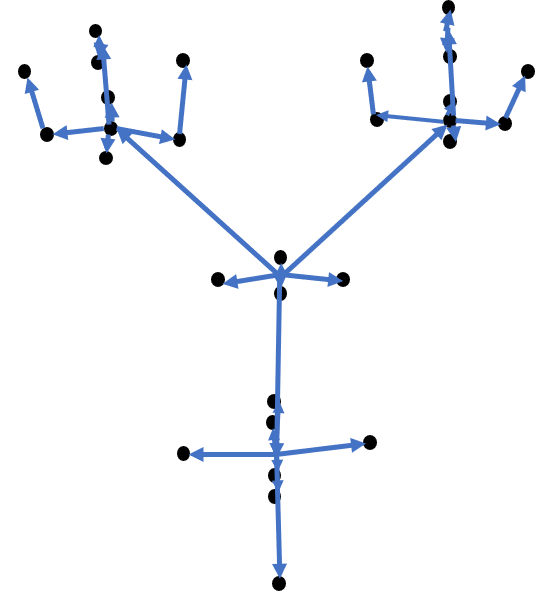}\hspace{25pt}\includegraphics[scale=0.21]{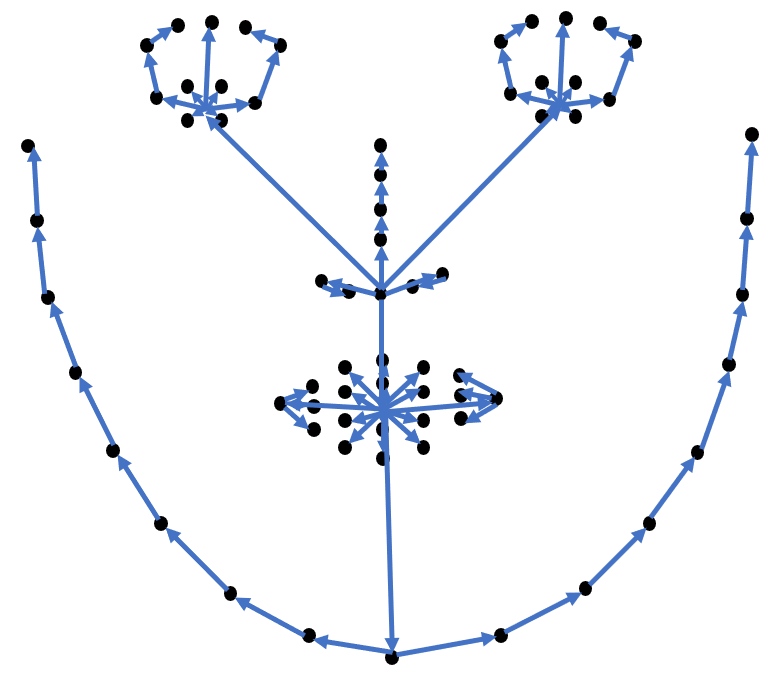}
\caption{\small The proposed extension of the dendritic structure from Figure \ref{fig:outline} generalizing to other datasets (COFW and 300W) each with different number of points.}
\label{fig:tree_extension}
\end{figure}

Each branch of PCD-CNN is designed according to the proposed Squeezenet-Deconv networks shown in Figure \ref{fig:deconv}. Due to fewer parameters in the Squeezent-Deconv, we hypothesize limited generalization capacity of the deconvolution network. By means of experiments, we show that effective training methods such as Mask-Softmax and Hard sample mining improves the performance of PCD-CNN by a large margin as a result of better generalization capacity.   

\textbf{Mask-Softmax Loss}:
To train the network, the localization of fiducial keypoints is formulated as a classification problem. The label for an input image of size $h\times w\times 3$ is a label tensor of same size as the image with $N+1$ channels, where $N$ is the number of keypoints. The first $N$ channels represent the location of each keypoint whereas the last channel represents the background. Each pixel is assigned a class label with invisible points being labeled as background. The objective is to minimize the following loss function: 
\begin{equation}
L_{0}(\Vec{p},\Vec{g}) = \sum_{i=1}^{h}\sum_{j=1}^{w}m(i,j)\sum_{k=1}^{N+1}g_{k}(i,j)log\bigg(\frac{e^{p_{k}(i,j)}}{\sum_{l}e^{p_{l}(i,j)}}\bigg)
\label{eq3}
\end{equation} 
where $k\in\{1,2\ldots N\}$ is the class index and $g_{k}(i,j)$ represents the ground truth at location $(i,j)$. $p_{l}(i,j)$ is the score obtained for location $(i,j)$ after forward pass through the network. Since the number of negative examples is orders of magnitudes larger than the positives, we design a strategic mask $m(i,j)$ which selects all the positive pixel samples, and keeps only $50\%$ of the 4-neighborhood pixels and $0.025\%$ of the negative background samples by random selection. During backward pass, the gradients are weighed accordingly. We experimentally show the effect of using Mask-Softmax Loss by training two separate PCD-CNN;  with and without the Mask-Softmax Loss; trained under identical training policies(Table \ref{mask_softmax_table}) . \\ 
\label{hard_mine}
\textbf{Hard Sample Mining}: \cite{c4} by Kabkab et al. showed that effective sampling of data improves the classification performance of the network. Following \cite{c4}, we use an offline hard sample mining procedure to train the proposed PCD-CNN. The histogram of error on the training data is plotted after the network is trained for 10 epochs by random sampling (refer supplementary material). We denote the mode of the distribution as $C$, and categorize all the training samples producing errors larger than $C$ as hard samples. Next we retrain the proposed PCD-CNN with hard and easy samples, sampled at the respective proportion. This effectively results in retraining the network by reusing the hard samples. Table \ref{hardmining_table} shows that such hard sample mining improves the performance of PCD-CNN (with fewer parameters) by a large margin.  
\begin{table}[]
\begin{subtable}[h]{0.5\textwidth}
\centering
\begin{tabular}{|l|l|}
\hline
\textbf{Method} & \textbf{Normalised Error} \\
\hline \hline 
Without Hard Mining & 2.85 \\
With Hard Mining & 2.49 \\
\hline
\end{tabular}
\caption{}
\label{hardmining_table}
\end{subtable}
\begin{subtable}[h]{0.5\textwidth}
\centering
\begin{tabular}{|l|l|}
\hline
\textbf{Method} & \textbf{Normalised Error} \\
\hline \hline 
Less Filters+Hard Mining & 2.49 \\
More Filters+Hard Mining & 2.40 \\
\hline
\end{tabular}
\caption{}
\label{more_filters_table}
\end{subtable}
\caption{\small Root mean square error normalized by bounding box calculated on the AFLW validation set following PIFA protocol. (a) depicts the effect of offline hard sample mining. (b) shows the effect of offline hard-mining and quadrupling the number of deconvolution filters.}
\end{table}

In the next set of experiments, we train PCD-CNN by increasing the number of deconvolution filters to 128 and 64 in each deconvolution network. We follow the same strategy of Mask-Softmax and hard sample mining to train this network.  Unsurprisingly, we see an improvement in performance for the task of keypoint localization (Table \ref{more_filters_table}), although, increasing the number of deconvolution filters leads to slower run time of $11$FPS as opposed to $20$FPS. 
\section{Experiments}
We select four different datasets with different characteristics to train and evaluate the proposed two stage PCD-CNN. 

\textbf{AFLW}\cite{tugraz:icg:lrs:koestinger11b}and \textbf{AFW}\cite{AFW_dataset_CVPR2012} are two \textit{difficult} datatsets which comprises of images in extreme pose, expression and occlusion. AFLW consists of $24,386$ in-the-wild faces (obtained from \textit{Flickr}) with head pose ranging from $0\degree$ to $120\degree$ for yaw and upto $90\degree$ for pitch and roll. AFLW provides at most 21 points for each face. It excludes coordinates for invisible landmarks and in our method such invisible points are labelled as background. For AFLW we follow the PIFA protocol; i.e. the test set is divided into three groups corresponding to three pose groups with equal number of images in each group.

AFW which is a popular benchmark for the evaluation of face alignment algorithms, consisting of 468 in-the-wild faces (also obtained from Flickr) with yaw up to $90\degree$. The images are diverse in terms of pose, expression and illumination and was considered the most difficult publicly available dataset, until AFLW. The number of visible points varies depending on the pose and occlusion with a maximum of 6 points per face image. We use AFW only for evaluation purposes. 

A \textit{medium} pose dataset from the popular \textbf{300W} face alignment competition\cite{6755925}. The dataset consists of re-annotated five existing datasets with 68 landmarks: iBug, LFPW, AFW, HELEN and XM2VTS. We follow the work \cite{Zhu_2015_CVPR} to use $3,148$ images for training and 689 images for testing. The testing dataset is split into three parts: common subset (554 images), challenging subset (135 images) and the full set (689 images).

Another dataset showing extreme cases of external and internal object \textit{occlusion}; \textbf{COFW}\cite{7410774}. COFW is the most challenging dataset that is designed to depict faces in real-world conditions with partial occlusions \cite{10.1109/ICCV.2013.191}. The face images show large variations in shape and occlusions due to differences in pose, expression, hairstyle, use of accessories or interactions with other objects. All 1,007 images were annotated using the same 29 landmarks as in the LFPW dataset, with their individual visibilities. The training set includes 845 LFPW faces + 500 COFW faces, that is 1,345 images in total. The remaining 507 COFW faces are used for testing. 

\textbf{Evaluation Metric}: Following  most  previous  works, we obtain the error for each test sample via averaging normalized errors for all annotated landmarks. We illustrate our results with mean error over all samples, or via Cumulative Error Distribution (CED) curve. For AFLW and AFW, the obtained error is normalized by the ground truth bounding box size over all visible points whereas for 300W and COFW, error is normalized by the inter-occular distance. Wherever applicable NME stands for Normalized Mean Error. 

\textbf{Training}: The PCD-CNN was first trained using the AFLW training set which was augmented by random cropping, flipping and rotation. The network was trained for $10$ epochs where the learning rate starting from $0.01$ was dropped every $3$ epochs. Keeping the weights of PCD-CNN fixed, the auxiliary network for fine grained classifcation was trained for another 10 epochs using the hard mining strategy explained in section \ref{hard_mine}. PoseNet was kept frozen while training the network for COFW and 300W datasets. All the experiments including training and testing were performed using the Caffe\cite{jia2014caffe} framework and Nvidia TITAN-X GPUs and p6000 GPUs. Being a non-iterative and single shot keypoint prediction method, our method is fast and can process \textbf{20} frames per second on 1 GPU only in batch mode. (Refer to supplementary material for more training details)
\begin{table}[h]
\begin{subtable}[h]{0.5\textwidth}
\centering
\begin{tabular}{|l|l|l|}
\hline
 & {\textbf{AFLW}} & {\textbf{AFW}} \\ 
\hline \hline
{{\textbf{Method}}} & {{\textbf{NME}}}  & {{\textbf{NME}}}\\ \hline \hline
TSPM \cite{AFW_dataset_CVPR2012}   & -     & 11.09   \\           
CDM \cite{Xiang_iccv_2013}         & 12.44    & 9.13          \\
RCPR \cite{10.1109/ICCV.2013.191}        & 7.85     &-          \\
ESR \cite{DBLP:journals/ijcv/CaoWWS14}         & 8.24    & -          \\
PIFA \cite{pifa}        & 6.8          & 9.42      \\
3DDFA \cite{DBLP:journals/corr/ZhuLLSL15}       & 5.32      &-         \\
LPFA-3D \cite{lfa3d}     & 4.72  & 7.43             \\
EMRT \cite{DBLP:journals/corr/ZhuLLT15}     & 4.01    & 3.55            \\ 
Hyperface \cite{DBLP:journals/corr/RanjanPC16}  & 4.26     &-          \\
Rec Enc-Dec\cite{recdec}  & \textgreater6        & -        \\ 
PIFAS\cite{pose-invariant-face-alignment-with-a-single-cnn} & 4.45 & 6.27 \\
FRTFA\cite{Bhagavatula_2017_ICCV} & 4.23 &- \\
CALE\cite{BMVC2016_86} & 2.63 & - \\
KEPLER\cite{7961750} & 2.98 & 3.01 \\
Binary-CNN\cite{Bulat_2017_ICCV} & 2.85 &- \\ \hline \hline
\textbf{PCD-CNN(Fast) Pre Test Aug}       & \textbf{2.85}   & \textbf{2.80}    \\
\textbf{PCD-CNN(Fast) Post Test Aug}       & \textbf{2.81}   & \textbf{2.66}    \\
\textbf{PCD-CNN(C+C) Pre Test Aug}       & \textbf{2.49}   & \textbf{2.52}    \\  
\textbf{PCD-CNN(C+C+more filters)}       & \textbf{2.40}   & \textbf{2.47}    \\ 
\textbf{PCD-CNN(C+C) Post Test Aug (Best)}       & \textbf{2.40}   & \textbf{2.36}    \\ 
\hline
\end{tabular}
\caption{}
\label{aflw_table}
\end{subtable}
\begin{subtable}[h]{0.5\textwidth}
\centering
\begin{tabular}{|l|l|l|l|l|}
\hline
\textbf{Method}                  & \textbf{{[}0,30{]}}    & \textbf{{[}30,60{]}}      & \textbf{{[}60,90{]}}   & \textbf{Mean}          \\ \hline \hline
HyperFace\cite{DBLP:journals/corr/RanjanPC16}               & 3.93          & 4.14          & 4.71          & 4.26          \\
AIO\cite{7961718}                     & 2.84          & 2.94          & 3.09          & 2.96          \\
Binary-CNN\cite{Bulat_2017_ICCV}           & 2.77          & 2.86          & 2.90          & 2.85          \\ \hline \hline
\textbf{PCD-CNN(C+C)} & \textbf{2.33} & \textbf{2.60} & \textbf{2.64} & \textbf{2.49} \\
\hline
\end{tabular}
\caption{}
\label{anglewise}
\end{subtable}
\caption{\small Comparison with previous methods on (a) AFLW-PIFA test set and AFW test set. (b) AFLW-PIFA categorized by absolute yaw angles. In (a) C+C stands for classification+classification. For AFLW, numbers for other methods are taken from respective papers following the PIFA protocol. For AFW, the numbers are taken from respective published works following the protocol of \cite{AFW_dataset_CVPR2012}. The numbers represent the normalized mean error.}
\end{table}

\begin{table}[h]
\begin{subtable}[h]{0.5\textwidth}
\centering
\begin{tabular}{|l|l|l|l|} 
\hline
\textbf{Method}  & \textbf{Common} & \textbf{Challenge} & \textbf{Full} \\
\hline \hline
RCPR\cite{10.1109/ICCV.2013.191}             & 6.18                & 17.26             & 8.35                     \\
SDM\cite{XiongD13}              & 5.57                & 15.40             & 7.52                     \\
ESR\cite{DBLP:journals/ijcv/CaoWWS14}              & 5.28                & 17.00             & 7.58                     \\
CFAN\cite{CFAN}             & 5.50                & 16.78             & 7.69                     \\
LBF\cite{DBLP:conf/cvpr/RenCW014}             & 4.95                & 11.98             & 6.32                     \\
CFSS\cite{Zhu_2015_CVPR}             & 4.73                & 9.98              & 5.76                     \\
TCDCN\cite{DBLP:conf/eccv/ZhangLLT14}           & 4.80                & 8.60              & 5.54                     \\
DDN\cite{DBLP:journals/corr/YuZC16}             & -                   & -                 & 5.59                     \\
MDM\cite{trigeorgis2016mnemonic}              & 4.83                & 10.14             & 5.88                     \\
TSR\cite{Lv_2017_CVPR}              & 4.36                & \textbf{7.56}     & 4.99                     \\ \hline \hline
\textbf{PCD-CNN} & \textbf{3.67}       & 7.62              & \textbf{4.44}  \\         
\hline
\end{tabular}
\caption{}
\label{ibug_res}
\end{subtable}
\begin{subtable}[h]{0.5\textwidth}
\centering
\begin{tabular}{|l|l|l|}
\hline
\textbf{Method}  & \textbf{NME}  & \textbf{Failure Rate} \\
\hline \hline
RCPR\cite{10.1109/ICCV.2013.191}              & 8.5           & 20\%                  \\
OFA\cite{Zhang_2016_CVPR}              & 6.46          & -                     \\
HPM\cite{6909641}              & 8.48          & 6.99\%                \\
ERCLM\cite{DBLP:journals/corr/BoddetiRSOK17}            & 6.49          & 6.3\%                 \\
RPP\cite{7084187}              & 7.52          & 16.2\%                \\ 
Human\cite{10.1109/ICCV.2013.191} & 5.6 & 0\% \\ \hline \hline
\textbf{PCD-CNN Pre Test Aug} & \textbf{6.02} & \textbf{4.53\%}  \\ 
\textbf{PCD-CNN Post Test Aug} & \textbf{5.77} & \textbf{3.73\%}  \\ 
\hline   
\end{tabular}
\caption{}
\label{cofw_table}
\end{subtable}
\caption{\small Comparison of the proposed method with other state-of-the-art methods on (a) 300W dataset (b) COFW testset. The NMEs for comparison on 300W dataset are taken from the Table 3 of \cite{Lv_2017_CVPR}.}
\end{table}

\subsection{Results}
\begin{figure*}[h]
    \centering
    \begin{subfigure}[htp!]{0.33\textwidth}
        \centering
        \includegraphics[height=4.5cm,width=\textwidth]{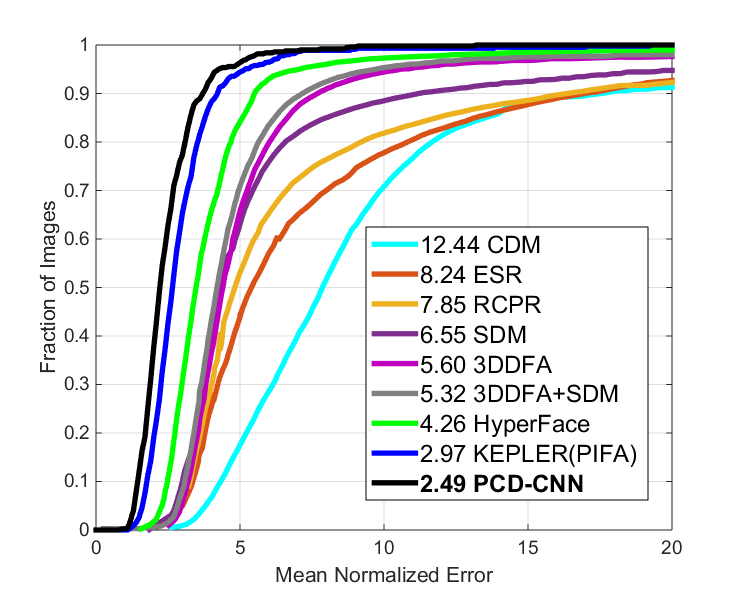}
        \caption{}
        \label{fig:aflw_res}
    \end{subfigure}  
    \begin{subfigure}[htp!]{0.33\textwidth}
        \centering
        \includegraphics[height=4.5cm,width=\textwidth]{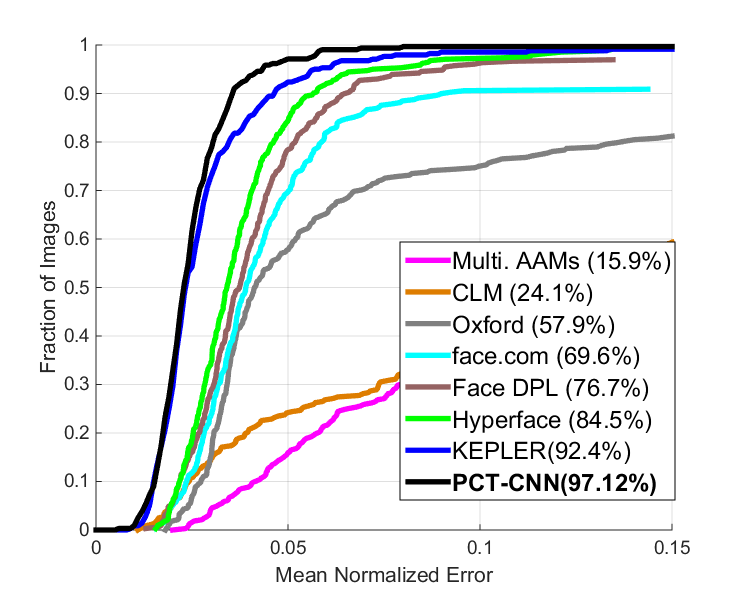}
        \caption{}
        \label{fig:afw_res}
    \end{subfigure}
    \begin{subfigure}[htp!]{0.33\textwidth}
		\centering
		\includegraphics[height=4.5cm,width=\textwidth]{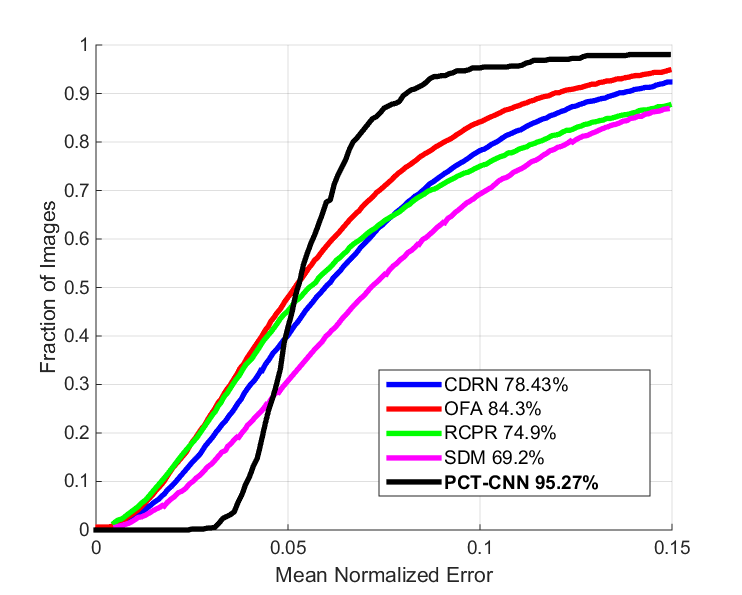}
		\caption{}
		\label{fig:cofw_res}
	\end{subfigure}
    \caption{\small Cumulative error distribution curves for landmark localization on AFLW, AFW and COFW dataset respectively. (a) Numbers in the legend represents mean error normalized by the face size. (b) Numbers in the legend are the fraction of testing faces that have average normalized error below $5\%$. (c) The numbers in the legend are the fraction of testing faces that have average normalized error below $10\%$. }
        \label{fig:res}
\end{figure*}

Table \ref{aflw_table} compares the performance of proposed method over other existing methods on AFLW-PIFA and AFW dataset. Table \ref{anglewise} compares the performance on AFLW-PIFA with respect to each pose group. Tables \ref{ibug_res} and \ref{cofw_table} compares the mean normalized error on the 300W and COFW datasets respectively. It is clear from the tables that while the proposed PCD-CNN performs comparable to previous state-of-the-art method\cite{Bulat_2017_ICCV}, the two stage PCD-CNN outperforms the state-of-the-art methods on all three datasets: AFLW, AFW and COFW by large margins. It is not surprising that increasing the number of deconvolution filters improves the performance on all the datasets. Figures \ref{fig:aflw_res}, \ref{fig:afw_res} and \ref{fig:cofw_res} show the cumulative error distribution for landmark localization in AFLW, AFW and COFW test sets. From the plots, we observe that the proposed PCD-CNN leads to a significant increase in the percentage of images with mean normalized error less than $5\%$. 
On AFW, fraction of images having an error of less than $15\degree$ for pose estimation is $87.22\%$ compared to $82\%$ in the recent work\cite{Hsu_2015_ICCV}. On COFW dataset, the NME reduces to $6.02$ (close human performance of $5.6$) bringing down the failure rate to $4.53\%$. PCD-CNN achieves a higher recall of $44.7\%$ at the precision of $80\%$ as opposed to RCPR's\cite{10.1109/ICCV.2013.191} $38.2\%$. (refer to the supplementary material for more results.)

\textbf{Improvement in localization by augmentation during testing :}
For a fair evaluation, we compare with the previous state-of-the-art methods with and without augmentation during testing. In the next set of experiments along with the test image, we also pass the flipped version of it and the final output is taken as the mean of the two outputs. With experimentation we observe that data augmentation while testing also improves the localization performance. While on AFLW-PIFA the error rate of $2.40$ is achieved, the effect of test set augmentation is more prominent in AFW dataset, where the error rate of $2.36$ is achieved. Similarly, on 300W (challenging) error rate drops to $7.17$ from $7.62$ as a result of test set augmentation. On COFW, error rate and failure rate of $5.77$ and $3.73\%$ respectively are achieved as the best results. 

Figure \ref{fig:qualitative} shows some of the difficult images and the predicted visible keypoints on the four datasets. We also achieve state of the art results on the performance of auxiliary tasks, such as pose estimation on AFW and occlusion prediction on COFW dataset. 
\begin{figure*}[h]
 \centering
\includegraphics[width=1.8cm,height=1.8cm]{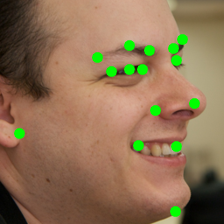}\hspace{3pt}\includegraphics[width=1.8cm,height=1.8cm]{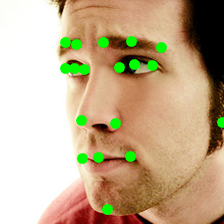}\hspace{3pt}\includegraphics[width=1.8cm,height=1.8cm]{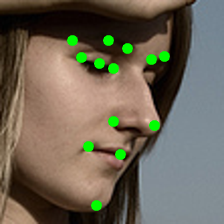}\hspace{3pt}\includegraphics[width=1.8cm,height=1.8cm]{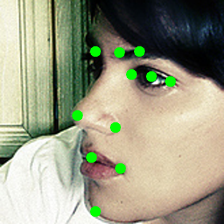}\hspace{3pt}\includegraphics[width=1.8cm,height=1.8cm]{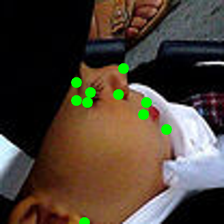}\hspace{3pt}\includegraphics[width=1.8cm,height=1.8cm]{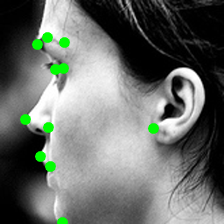}\hspace{3pt}\includegraphics[width=1.8cm,height=1.8cm]{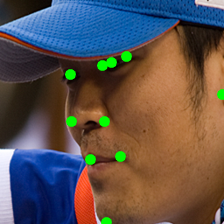}\\
\includegraphics[width=1.8cm,height=1.8cm]{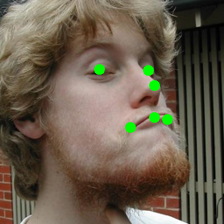}\hspace{3pt}\includegraphics[width=1.8cm,height=1.8cm]{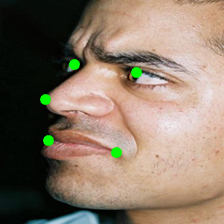}\hspace{3pt}\includegraphics[width=1.8cm,height=1.8cm]{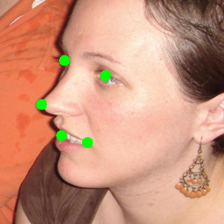}\hspace{3pt}\includegraphics[width=1.8cm,height=1.8cm]{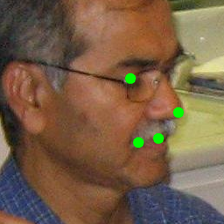}\hspace{3pt}\includegraphics[width=1.8cm,height=1.8cm]{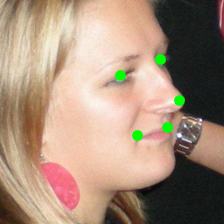}\hspace{3pt}\includegraphics[width=1.8cm,height=1.8cm]{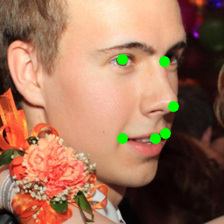}\hspace{3pt}\includegraphics[width=1.8cm,height=1.8cm]{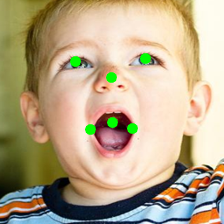}\\
\includegraphics[width=1.8cm,height=1.8cm]{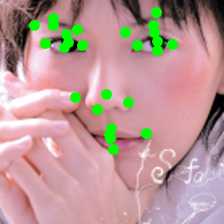}\hspace{3pt}\includegraphics[width=1.8cm,height=1.8cm]{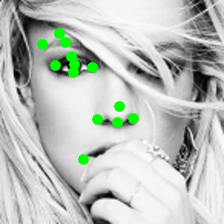}\hspace{3pt}\includegraphics[width=1.8cm,height=1.8cm]{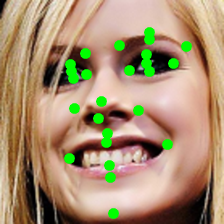}\hspace{3pt}\includegraphics[width=1.8cm,height=1.8cm]{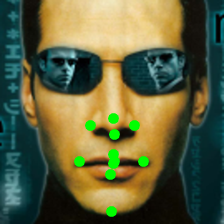}\hspace{3pt}\includegraphics[width=1.8cm,height=1.8cm]{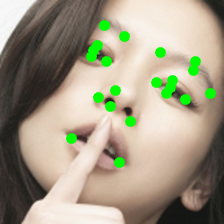}\hspace{3pt}\includegraphics[width=1.8cm,height=1.8cm]{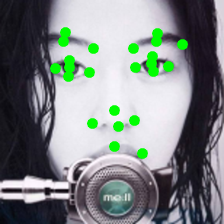}\hspace{3pt}\includegraphics[width=1.8cm,height=1.8cm]{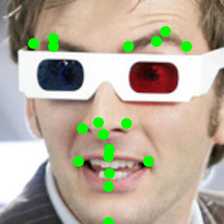}\\
\includegraphics[width=1.8cm,height=1.8cm]{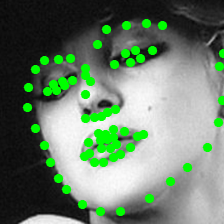}\hspace{3pt}\includegraphics[width=1.8cm,height=1.8cm]{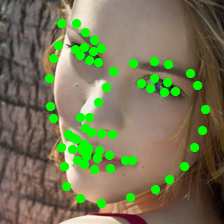}\hspace{3pt}\includegraphics[width=1.8cm,height=1.8cm]{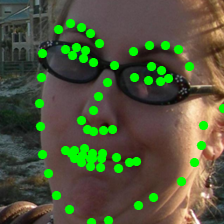}\hspace{3pt}\includegraphics[width=1.8cm,height=1.8cm]{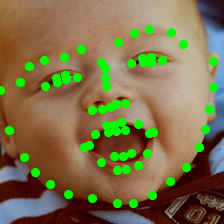}\hspace{3pt}\includegraphics[width=1.8cm,height=1.8cm]{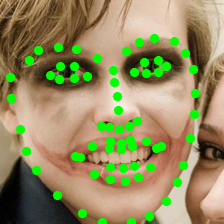}\hspace{3pt}\includegraphics[width=1.8cm,height=1.8cm]{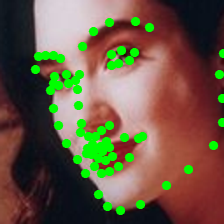}\hspace{3pt}\includegraphics[width=1.8cm,height=1.8cm]{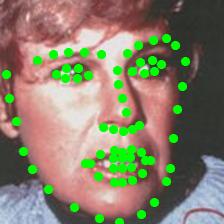}
\caption{\small Qualitative results generated from the proposed method. The green dots represent the predicted points. Each row shows some of the difficult samples from AFLW, AFW, COFW, and 300W respectively with all the visible predicted points.}
\label{fig:qualitative}
\end{figure*}
\section{Conclusion and Future Work}
In this paper, we present a dendritic CNN which processes images at full scale looking at the images globally and capturing local interactions through convolutions. The proposed PCD-CNN is able to precisely localize landmark points on unconstrained faces without using any 3D morphable models. We also demonstrate that disentangling pose by conditioning on it can influence the localization of landmark points by reducing the mean pixel error by a large margin. Due to effective design choices made, the proposed model is not limited to yield a fixed number of points and can be extended to other datasets with different protocols. With the help of ablative studies, impact of effective training of the convolutional network by using sampling strategies such as Mask-Softmax and hard instance sampling is shown. Using smaller and fewer convolution filters, the proposed network is able to process images close to real-time and can be deployed in a real life scenario. The proposed method can be easily extended to 3D dense face alignment and other tasks, which we plan to pursue in the future. 
\section{Acknowledgment}
This research is based upon work supported by the Office of the Director of National Intelligence (ODNI), Intelligence Advanced Research Projects Activity (IARPA), via IARPA R\&D Contract No. 2014-14071600012. The views and conclusions contained herein are those of the authors and should not be interpreted as necessarily representing the official policies or endorsements, either expressed or implied, of the ODNI, IARPA, or the U.S. Government. The U.S. Government is authorized to reproduce and distribute reprints for Governmental purposes notwithstanding any copyright annotation thereon.
We also thank our colleagues for all the discussion sessions.  

{\small
\bibliographystyle{ieee}
\bibliography{ref}
}

\clearpage
\section*{Disentangling 3D-Pose in A Dendritic CNN \\for Unconstrained 2D-Face Alignment - Supplementary}

\section{Effect of Pose Disentaglement}
Next, we also perform an experiment to observe the effect of 3D pose conditioning on the second auxiliary network designed for fine grained localization. Table \ref{pose_condition_table_aux} shows the effect of disentangling pose by conditioning, when the auxiliary conv-deconv network does not receive information from the PoseNet. 

\begin{table}[htp]
\centering
\begin{tabular}{|l|l|}
\hline
\textbf{Method} & \textbf{NME} \\
\hline \hline 
PCD-CNN + Auxiliary Network & 2.99 \\
PCD-CNN + Pose Conditioned Auxiliary Network & 2.49 \\
\hline
\end{tabular}
\caption{Mean square error normalized by bounding box calculated on AFLW test set following PIFA protocol. When PCD-CNN and fine-grained localization network both are conditioned on pose yields lower error rate.}
\label{pose_condition_table_aux}
\end{table}

\section{Magnified version of the Tree}
One expects to receive information from all other keypoints in order to optimize the features at a specific keypoint. However, this has two drawbacks: First, to model the interaction between keypoints lying far away such as `eye corner' and `chin', convolution kernels with larger size have to be introduced. This leads to increase in the number of parameters. Secondly, relationships between some keypoints are unstable, such as `left eye corner' and `right eye corner'. In a profile face image one of the points may not be visible and passing information between those two keypoints may lead to erroneous results. Hence, convolution kernels are learned at the size of $14\times14$ which ensures keypoints which are closer and have stable relationships to be connected together.

\begin{figure*}[htp!]
    \centering
    \begin{subfigure}[htp!]{0.33\textwidth}
        \centering
        \includegraphics[height=5.5cm,width=1\textwidth]{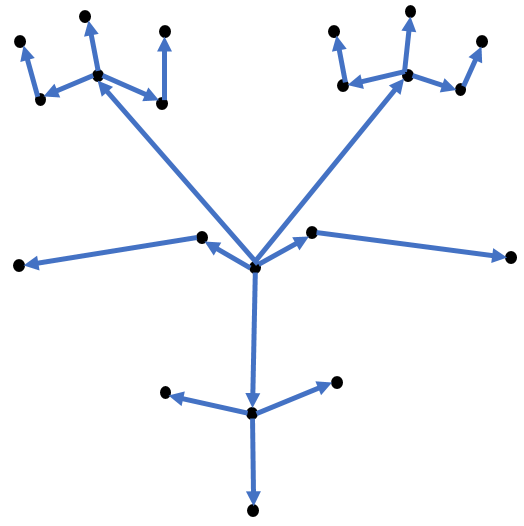}
        \caption{}
        \label{fig:aflw_tree}
    \end{subfigure}\hfill 
    \begin{subfigure}[htp!]{0.33\textwidth}
        \centering
        \includegraphics[height=5.5cm,width=1\textwidth]{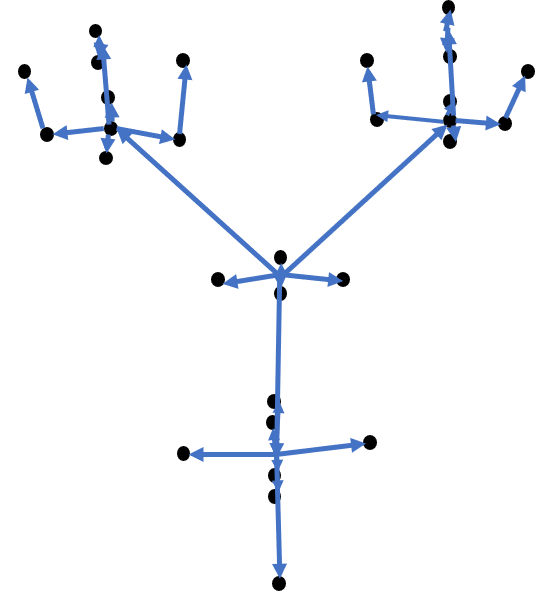}
        \caption{}
        \label{fig:cofw_tree}
    \end{subfigure}\hfill
    \begin{subfigure}[htp!]{0.33\textwidth}
        \centering
        \includegraphics[height=5.5cm,width=1\textwidth]{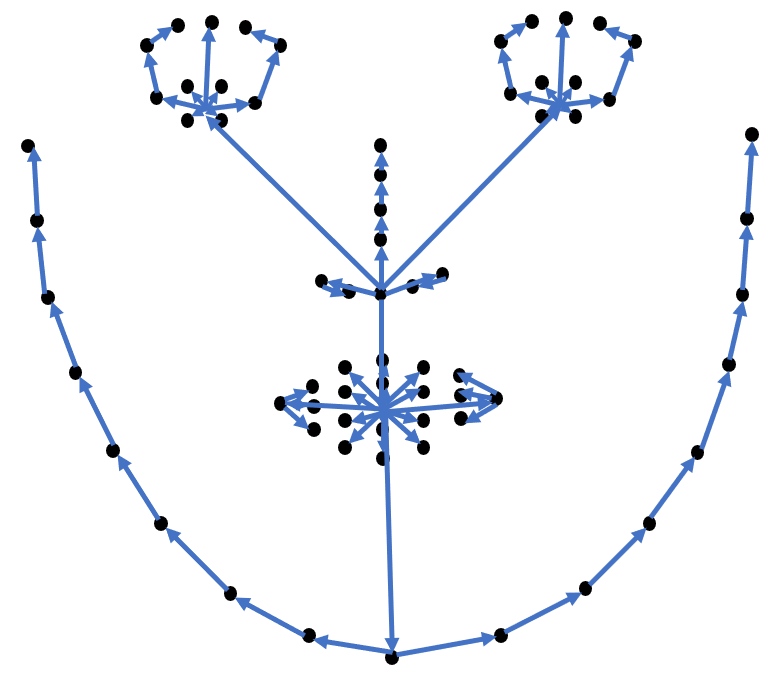}
        \caption{}
        \label{fig:300w_tree}
    \end{subfigure}
    \caption{The proposed extension of the dendritic structure from Figure 1 of the main paper, generalizing to other datasets with variable number of points.}
        \label{fig:tree_extension}
\end{figure*}

We also describe the process of extending the proposed dendritic structure of facial landmarks to other datasets with variable number of landmark points. Figure \ref{fig:aflw_tree} shows the tree structure of the 21 landmark points compatible with the AFLW dataset. In figure \ref{fig:cofw_tree} and \ref{fig:300w_tree} the number of points is increased to 29 and 68 respectively compatible with COFW and 300W datasets. We wish to keep the structure of the facial landmarks intact while increasing the number of landmark points. For this, we make use of the network surgery. First, the number of deconvolution filters in the penultimate and ultimate deconvolution layers is increased to 128 and 64 respectively. Next $1\times 1$ convolutions are used to obtain desire number of outputs, which is then sliced and concatenated in order for loss computation. For instance, eye center points is split into $4$ landmark points in the case of COFW and 300W datasets, and ear corner points are dropped. An advantage of network surgery is that, it leads to yielding a variable number of landmark points with minimal increase in parameters while keeping the face structure intact.  

\section{Training Details}
KeypointNet and PoseNet described in section $3$ are designed based on the SqueezeNet architecture, attributing its lower parameter count. The proposed PCD-CNN was first trained using AFLW training set, where Mask-Softmax is used for keypoints and Euclidean Loss for 3D pose estimation. Starting from the learning rate of $0.001$, the network was trained for $10$ epochs with momentum set to $0.95$. The learning rate was dropped by a factor of $10$ every $3$ epochs. While training PCD-CNN for COFW and 300W datasets, the convolution branch was initialized with the previously trained network, whereas the deconvolution branches were trained from scratch. Since, COFW and 300W datasets does not provide 3D pose ground truth, we leverage the previously trained PoseNet and freeze its weights. As shown in the section $3$ of the main paper, disentangling pose by conditioning improves the localization performance. 

\subsection{Training PCD-CNN for COFW}
This section covers the details of training for the COFW dataset. The PCD-CNN network was trained using the Mask Softmax and hard negative mining. The second auxiliary network was trained for the task of occlusion detection. According to the released details about the COFW dataset, around 23\% of the landmark points are invisible. Hence, to tackle the class imbalance problem between the visible and invisible points the following loss function was used. 
\begin{equation}
L(\Vec{p},\Vec{g}) = \sum_{i=1}^{29} (0.23*\mathbbm{1}_{g_{i}^{vis}=1} + 0.77*\mathbbm{1}_{g_{i}^{vis}=0})(p_{i}^{vis} - g_{i}^{vis})^{2} 
\end{equation}
where $\Vec{p},\Vec{g}$ are the vector of predicted and ground-truth visibilities. $p_{i}^{vis}$ and $g_{i}^{vis}$ are the values of the individual elements in the vectors of visibilities. The weighted loss function also balances the gradients back-propagated while loss calculation. 

\begin{figure}[htp!]
\centering
\includegraphics[height=9cm,width=0.5\textwidth]{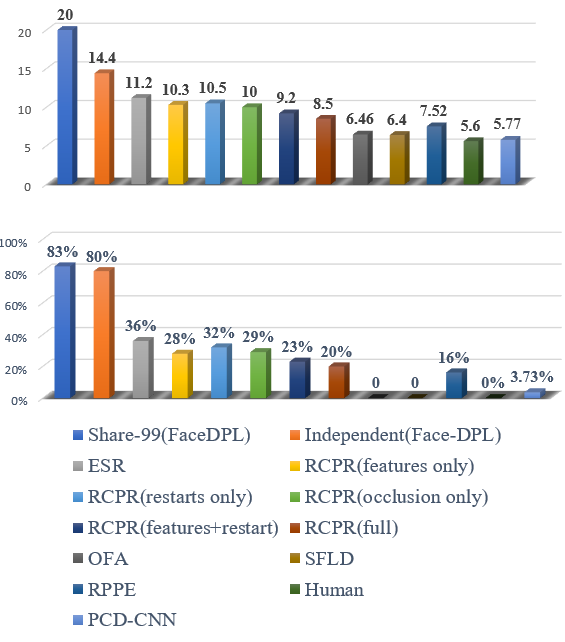}
\caption{Comparison of NME and failure rate over visible landmarks out of 29 landmarks from the COFW dataset.}
\label{fig:cofw_numbers}
\end{figure}

\begin{figure*}[htp!]
    \centering
    \begin{subfigure}[htp!]{0.35\textwidth}
        \centering
        \includegraphics[height=5.5cm,width=\textwidth]{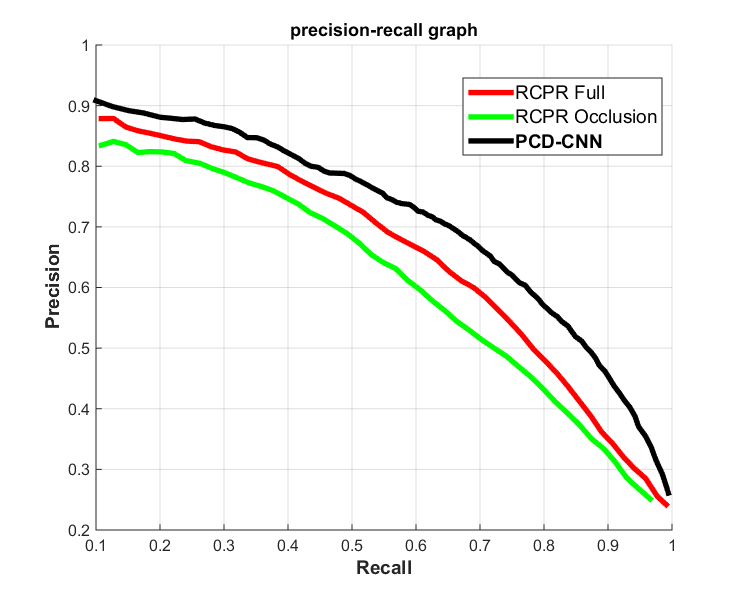}
		\caption{}
		\label{fig:cofw_vis}
    \end{subfigure} 
    \begin{subfigure}[htp!]{0.35\textwidth}
        \centering
        \includegraphics[height=5.5cm,width=\textwidth]{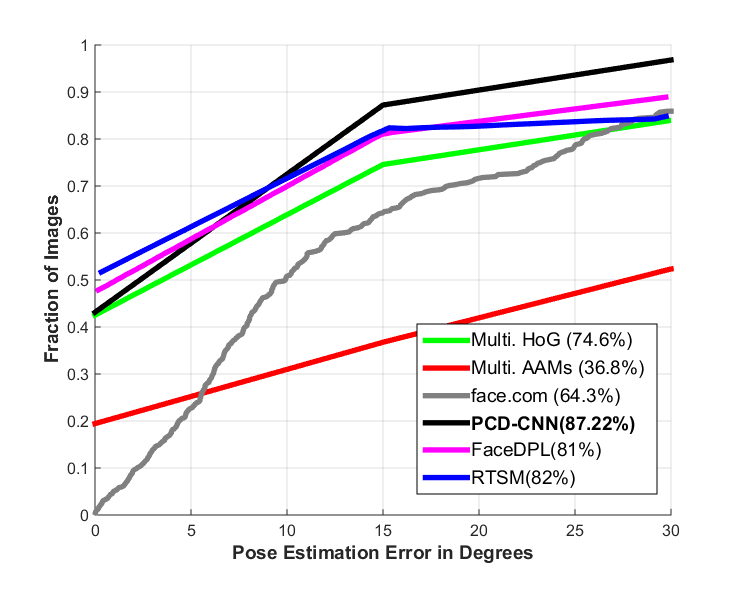}
		\caption{}
		\label{fig:afw_pose}
    \end{subfigure} 
    \begin{subfigure}[htp!]{0.35\textwidth}
        \centering
        \includegraphics[height=5.5cm,width=\textwidth]{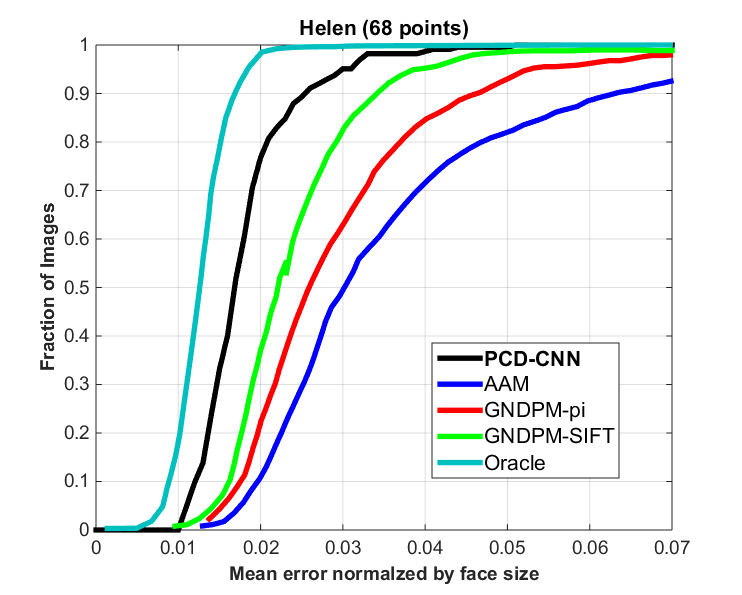}
        \caption{}
        \label{fig:helen}
    \end{subfigure} 
    \begin{subfigure}[htp!]{0.35\textwidth}
        \centering
        \includegraphics[height=5.5cm,width=\textwidth]{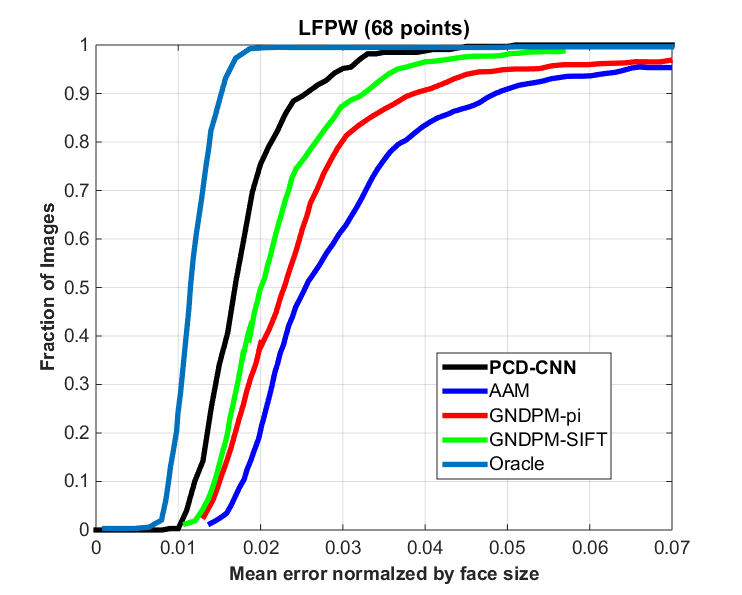}
        \caption{}
        \label{fig:lfpw}
    \end{subfigure}
    \caption{ (a) Precision Recall for the occlusion detection on the COFW dataset. (b)Cumulative error distribution curves for pose estimation on AFW dataset. The numbers in the legend are the percentage of faces that are labeled within $\pm15\degree$ error tolerance. Cumulative Error Distribution curve for (c) Helen (d) LFPW, when the average error is normalized by the bounding box size.}
        \label{fig:300w}
\end{figure*}

Figure \ref{fig:cofw_numbers} shows the failure rate and error rate on the COFW dataset. The failure rate on the COFW dataset drops to $4.53\%$ bringing down the error rate to $6.02$. When testing with the augmented images the error rate further drops to $5.77$ bringing it closer to human performance $5.6$. Figure \ref{fig:cofw_vis} shows the precision recall curve for the task of occlusion detection on the COFW dataset. PCD-CNN achieves a significantly higher recall of $44.7\%$ at the precision of $80\%$ as opposed to RCPR's\cite{10.1109/ICCV.2013.191} $38.2\%$. 

\section{Hard mining}
\begin{figure}[htp!]
    \centering
    \begin{subfigure}[htp!]{0.5\textwidth}
        \centering
        \includegraphics[height=3cm,width=\textwidth]{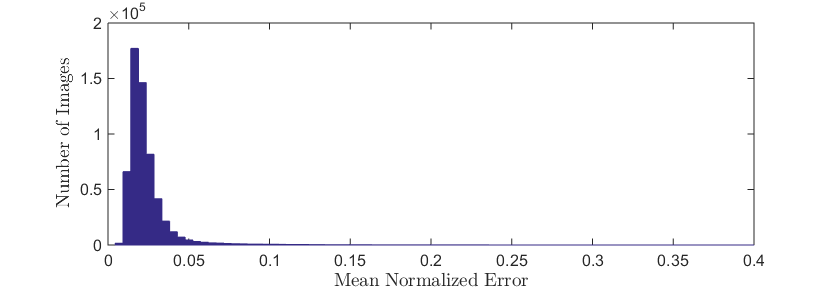}
        \caption{}
        \label{fig:outline_a}
    \end{subfigure} \\
    \begin{subfigure}[htp!]{0.5\textwidth}
        \centering
        \includegraphics[height=3cm,width=\textwidth]{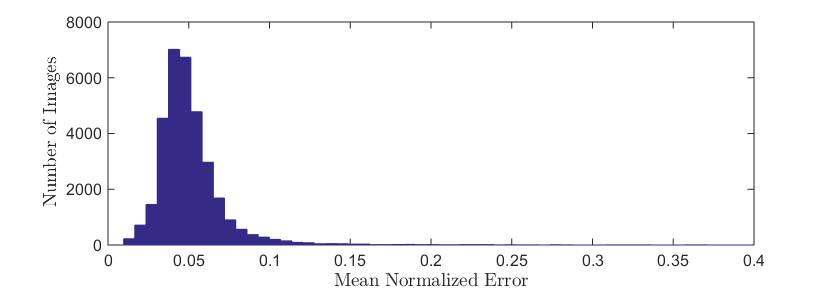}
        \caption{}
        \label{fig:outline_b}
    \end{subfigure}
    \caption{ Histogram of error, when evaluated on the training set of (a) AFLW  (b) COFW.}
        \label{fig:hard}
\end{figure}

Figure \ref{fig:hard} shows the distribution of average normalized error on the training sets of AFLW and COFW datasets. The error distributions were obtained upon evaluating the PCD-CNN network on the training set, after it is trained  with the whole dataset for 10 epochs. The dataset is partitioned into hard and easy samples after choosing the mode of the distribution as the threshold. Next, the network is trained again, by sampling equal number of images from both groups, which results in an effective reuse of the hard examples. 

\begin{figure*}[htp!]
 \centering
\includegraphics[width=2.5cm,height=2.5cm]{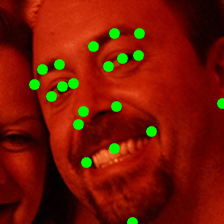}\includegraphics[width=2.5cm,height=2.5cm]{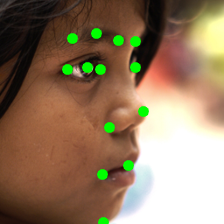}\includegraphics[width=2.5cm,height=2.5cm]{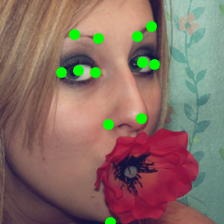}\includegraphics[width=2.5cm,height=2.5cm]{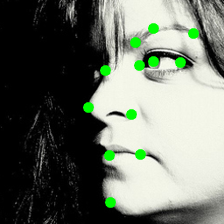}\includegraphics[width=2.5cm,height=2.5cm]{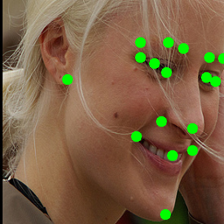}\includegraphics[width=2.5cm,height=2.5cm]{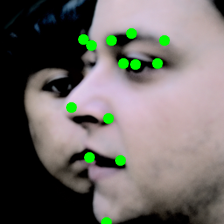}\\ \includegraphics[width=2.5cm,height=2.5cm]{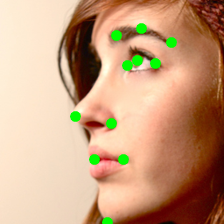}\includegraphics[width=2.5cm,height=2.5cm]{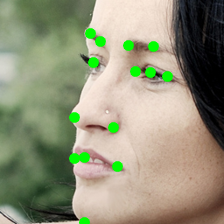}\includegraphics[width=2.5cm,height=2.5cm]{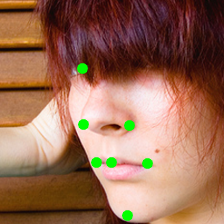}\includegraphics[width=2.5cm,height=2.5cm]{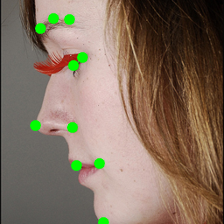}\includegraphics[width=2.5cm,height=2.5cm]{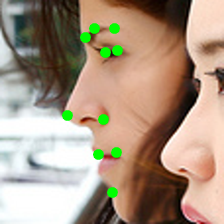}\includegraphics[width=2.5cm,height=2.5cm]{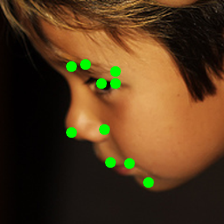}\\

\includegraphics[width=2.5cm,height=2.5cm]{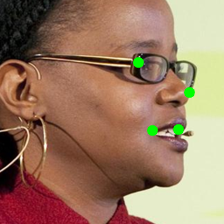}\includegraphics[width=2.5cm,height=2.5cm]{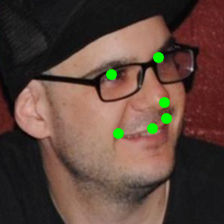}\includegraphics[width=2.5cm,height=2.5cm]{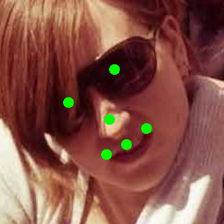}\includegraphics[width=2.5cm,height=2.5cm]{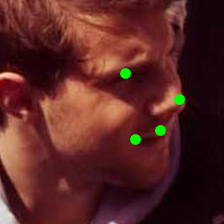}\includegraphics[width=2.5cm,height=2.5cm]{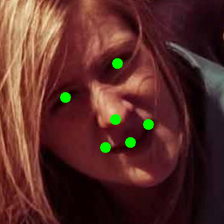}\includegraphics[width=2.5cm,height=2.5cm]{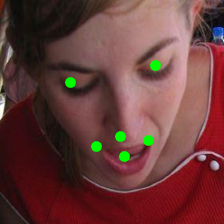}\\
\includegraphics[width=2.5cm,height=2.5cm]{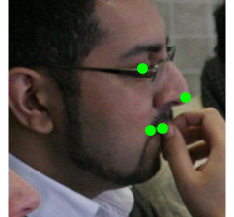}\includegraphics[width=2.5cm,height=2.5cm]{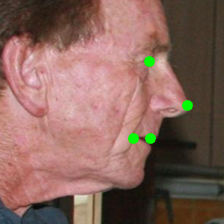}\includegraphics[width=2.5cm,height=2.5cm]{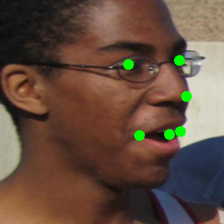}\includegraphics[width=2.5cm,height=2.5cm]{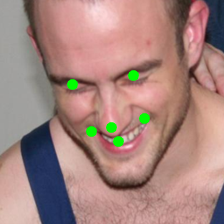}\includegraphics[width=2.5cm,height=2.5cm]{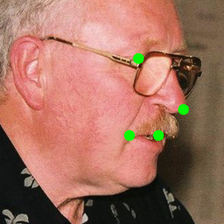}\includegraphics[width=2.5cm,height=2.5cm]{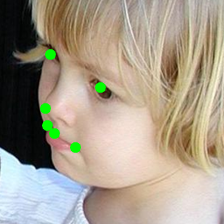}\\

\includegraphics[width=2.5cm,height=2.5cm]{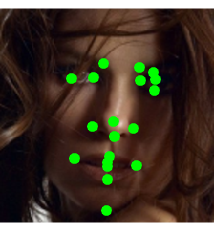}\includegraphics[width=2.5cm,height=2.5cm]{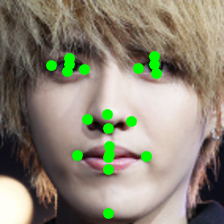}\includegraphics[width=2.5cm,height=2.5cm]{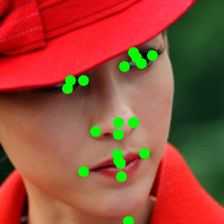}\includegraphics[width=2.5cm,height=2.5cm]{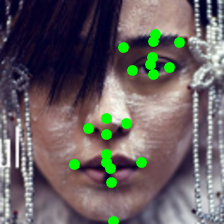}\includegraphics[width=2.5cm,height=2.5cm]{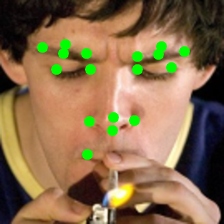}\includegraphics[width=2.5cm,height=2.5cm]{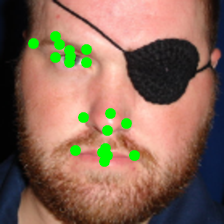}\\
\includegraphics[width=2.5cm,height=2.5cm]{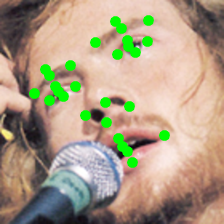}\includegraphics[width=2.5cm,height=2.5cm]{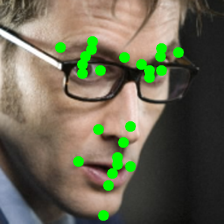}\includegraphics[width=2.5cm,height=2.5cm]{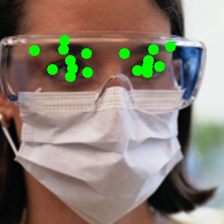}\includegraphics[width=2.5cm,height=2.5cm]{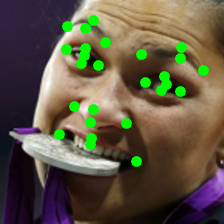}\includegraphics[width=2.5cm,height=2.5cm]{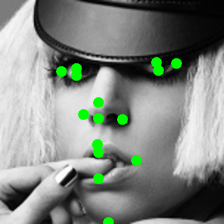}\includegraphics[width=2.5cm,height=2.5cm]{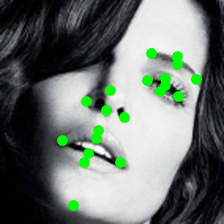}\\

\includegraphics[width=2.5cm,height=2.5cm]{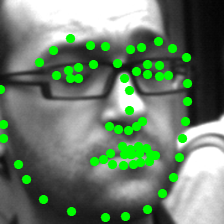}\includegraphics[width=2.5cm,height=2.5cm]{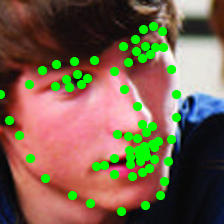}\includegraphics[width=2.5cm,height=2.5cm]{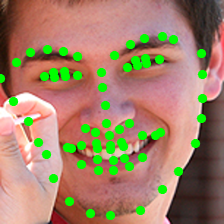}\includegraphics[width=2.5cm,height=2.5cm]{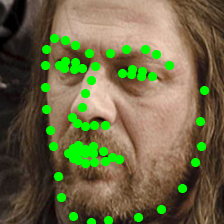}\includegraphics[width=2.5cm,height=2.5cm]{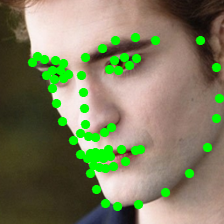}\includegraphics[width=2.5cm,height=2.5cm]{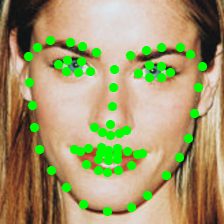}\\
\includegraphics[width=2.5cm,height=2.5cm]{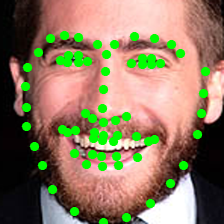}\includegraphics[width=2.5cm,height=2.5cm]{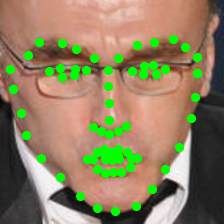}\includegraphics[width=2.5cm,height=2.5cm]{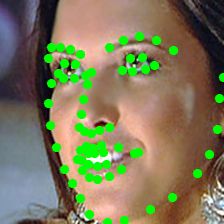}\includegraphics[width=2.5cm,height=2.5cm]{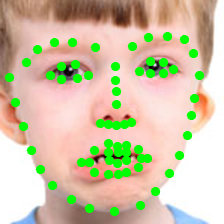}\includegraphics[width=2.5cm,height=2.5cm]{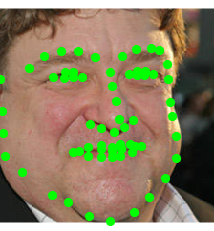}\includegraphics[width=2.5cm,height=2.5cm]{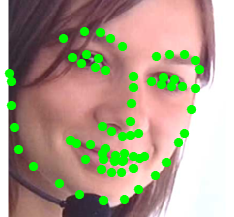}
\caption{Qualitative results generated from the proposed method. The green dots represent the predicted points. Every two rows show randomly selected samples from AFLW, AFW, COFW, and 300W respectively with all the visible predicted points.}
\label{fig:qualitative}
\end{figure*}
\section{More results on AFLW, AFW, LFPW and HELEN}

In this section, we show some more results obtained by the PCD-CNN on AFW, LFPW and Helen datasets. Figure \ref{fig:afw_pose} shows the cumulative error distribution curves for the prediction of face pose on AFW dataset. We observe that even though the primary objective of PCD-CNN is not pose prediction, it achieves state-of-the-art results when compared to recently published works Face-DPL\cite{AFW_dataset_CVPR2012},RTSM\cite{Hsu_2015_ICCV}. 

Figures \ref{fig:helen} and \ref{fig:lfpw} show the cumulative error distribution curve on LFPW and Helen datasets, when the average error is normalized by face size. PCD-CNN achieves significant improvement over the recent work of GNDPM\cite{6909635}. 

Figure \ref{fig:qualitative} shows some of the difficult test samples from AFLW, AFW, COFW and IBUG datasets respectively. 

\end{document}